\documentclass[runningheads]{llncs}

\usepackage{algorithmic,xcolor,amssymb,amsmath,float,graphicx,wasysym,marvosym,lipsum,enumitem,textcomp,latexsym,algorithm,xspace}

\newcommand{\kb}{KB\xspace}

\newcommand{\sdp}{\emph{SD-PIA}\xspace}

\begin{document}
\title{Augmenting Robot Knowledge Consultants\\ with Distributed Short Term Memory}
\titlerunning{Augmenting Robot Knowledge Consultants}
\authorrunning{Williams, Thielstrom, Krause, Oosterveld \& Scheutz}

\author{Tom Williams\inst{1}\and Ravenna Thielstrom\inst{2}\and Evan
  Krause\inst{2}\and Bradley Oosterveld\inst{2}\and Matthias Scheutz\inst{2}
}

\institute{Colorado School of Mines MIRRORLab, Golden, CO, USA\\
        \email{twilliams@mines.edu}, \url{mirrorlab.mines.edu}\and
Tufts University Human-Robot Interaction Lab,  Medford, MA, USA\\
        \email{\{firstname.lastname\}@tufts.edu}, \url{hrilab.tufts.edu}
}

\maketitle


\begin{abstract}

  Human-robot communication in situated environments involves a complex
  interplay between knowledge representations across a wide variety of
  modalities. Crucially, linguistic information must be associated
  with representations of objects, locations, people, and goals, which
  may be represented in very different ways. In previous work, we
  developed a Consultant Framework that facilitates modality-agnostic
  access to information distributed across a set of heterogeneously
  represented knowledge sources. 
  In this work, we draw inspiration from cognitive
  science to augment these distributed knowledge sources with Short
  Term Memory Buffers 
  to create an STM-augmented algorithm for referring expression
  generation. We then discuss the potential performance benefits of this
  approach 
  and insights from
  cognitive science that may inform future refinements in the design
  of our approach. 

\end{abstract}


%
%

\section{Introduction}

Social robots engaging in natural task-based dialogues with human
teammates must understand and generate natural
language expressions that refer to entities
such as people, locations, and objects~\cite{mavridis2015review,popescu1998reference}. These tasks, known as
\emph{reference resolution} and \emph{referring expression
  generation (REG)}, are particularly
challenging in realistic robotics applications due to the realities of
how knowledge is represented and distributed in modern robotic architectures.

We previously presented a \emph{Consultant Framework}~\cite{williams2017iib} that
allows a robot to use its distributed sources of knowledge during reference resolution~\cite{williams2016aaai}
and REG~\cite{williams2017inlg}, without requiring the language
processing system to understand how that knowledge is represented and accessed.
We've used this framework in
previous work to enable a modern take on the classic Incremental
Algorithm (IA)~\cite{dale1995computational} for REG, relaxing
several assumptions: that 
knowledge is certain, that knowledge is centrally stored, and that
a list of all properties known to hold for each known entity is
centrally available during REG. Our
Consultant Framework allows these assumptions to be relaxed, producing
a REG algorithm tailored to the realities
of robotic architectures.
Domain independence, however, comes at a computational
cost, especially for language generation. 

The IA requires, for each property that could be included,
consideration of whether it holds for the
to-be-described target and not for at least one
distractor. Under the assumptions of the IA, this can be performed via
set-membership checks on centrally available property sets. When
the assumption of such property sets is relaxed,
however, as is the case in the modified algorithm designed to leverage
our Consultant Framework, these considerations must instead be made through
queries to the Consultants responsible for the target and
distractors. The computational complexity of REG combined with the computational cost of these queries
results in a significant computational burden.

To address this computational burden, we propose(see also~\cite{williams2018mrhrcstmb}) an augmented
Consultant Framework that includes Consultant-Specific Short Term
Memory (STM) Buffers that cache a small number of properties recently
determined to hold for various entities.
We will begin by defining this augmented framework, and by describing how it
reduces the complexity of REG. We will
then discuss different possible assumptions that can be made in the
design of these STM Buffers, and discuss how these
different choices impact both efficiency and cognitive
plausibility. Next, we will discuss insights that can be gleaned from
psychological models of memory decay and forgetting and computational
caching strategies, and how these insights apply to the design of
these STM Buffers. Finally, we will discuss how these
Buffers can be exploited by processes beyond REG, and their potential relation to other cognitive models
maintained throughout the architecture.

\section{Augmented Framework}

Our previously presented Consultant Framework~\cite{williams2017inlg}
allows information about entities to be assessed when knowledge is
uncertain, heterogeneous, and
distributed, facilitating IA-inspired approaches to REG. Specifically, each   
Consultant $c$ facilitates access to one \kb $k$, and must
be capable of four functions: 
\begin{small}
\begin{enumerate}
\item providing a set $c_{domain}$ of atomic entities from $k$,
\item advertising a list $c_{constraints}$ of constraints that can
be assessed with respect to entities from $c_{domain}$, \emph{and
that is ordered by descending 
preference}.
\item assessing constraints from $c_{constraints}$ with respect to
entities from $c_{domain}$, and
\item adding, removing, or imposing constraints from $c_{constraints}$
on entities from $c_{domain}$.
\end{enumerate}
\end{small}
In this section, we define an \emph{(STM)-Augmented
  Consultant Framework} that adds an additional requirement:
\begin{small}
\begin{enumerate}
\setcounter{enumi}{4}
\item providing a list $c_{STM}$ of properties that hold for some
  entity from $c_{domain}$.
\end{enumerate}
\end{small}
Crucially, the properties returned through this capability do not need
to be all of the properties that hold for the target entity. A
Consultant may have a large number of properties that it could assess
for a given entity if need be, some of which might be very expensive
to compute. As such, the purpose of this capability is not to request
evaluation of all possible properties for the specified entity, but
rather to request the contents of a small cache of properties recently
determined to hold for the specified entity.

Models of Working Memory
suggest that humans maintain cached knowledge of
a small number of activated entities. While early  models of working
memory suggested that the size of working memory is bounded to a limited
\emph{number of chunks} (as in Miller's famous ``magical number'' of
seven, plus or minus two~\cite{miller1956magical}), more recent
models instead suggest that the size of working
memory is affected by the complexity of those chunks
~\cite{mathy2012s}. For
example, needing to maintain
multiple \emph{features} of a single entity may
detract
from the total number of maintainable entities, and
accordingly, the number 
of features maintainable for other
entities~\cite{alvarez2004capacity,taylor2017does,oberauer2016limits}. 
Moreover, recent research has suggested that humans may have different
resource limits for different \emph{types} of representations (e.g.,
visual vs. auditory~\cite{fougnie2015working}, or different types of
visual features~\cite{wang2017separate}) either due to the existence
of separate domain-specific cognitive
resources~\cite{baddeley1992working,logie2014visuo}
or do to decreased interference between disparate
representations~\cite{oberauer2012modeling}.

Drawing on these insights, the new capability required in the
\emph{STM-Augmented Consultant Framework} requires each Consultant to
maintain its own set of
\emph{features} currently remembered for the set of entities for which
it is responsible. This serves to allow fast access to a set of entity
properties likely to be relevant, in order to avoid the expensive
long-term memory queries that make processes such as referring
expression generation so expensive in the current Consultant
framework. 
In the next section we describe how our newly proposed framework can
be used during the course of REG.

\section{Algorithmic Approach}

We now present \sdp, the STM-Augmented variant of the 
Distributed, Probabilistic
IA~\cite{williams2017inlg}. We will describe the main
differences between \emph{DIST-PIA} and \sdp, the key difference being
our use of the properties stored in STM before  
performing LTM-Query intensive operations.  
%
While DIST-PIA crafted sub-descriptions through the use of a
single algorithm (DIST-PIA-H), \sdp begins by crafting an
initially (possibly partial) sub-description using the
\emph{SD-PIA-STM-H} algorithm, which utilizes only the properties found
in STM Buffers. If the sub-descriptions returned
through this algorithm are not fully discriminating, the partial
sub-description is augmented by passing the set of still-to-be-eliminated distractors to
\emph{SD-PIA-H}, which operates much the same as the original
\emph{DIST-PIA-H} algorithm.
  
The other main difference between \emph{DIST-PIA} and \sdp
is in the design of the \emph{SD-PIA-STM-H} function. Instead
of considering all properties advertised in the target's domain,
\emph{SD-PIA-STM-H} considers only the properties 
returned by querying that Consultant's STM buffer, requiring a single
query rather than $O(c_m^{\Lambda})$ queries. For each of these
already-known-to-hold and already-bound queries,
\emph{SD-PIA-STM-H} iteratively rebinds the query to refer to
each distractor $x$ rather than the target entity. For each re-bound
query, \emph{SD-PIA-STM-H} calls  a function \emph{stm-apply},
which checks whether that property holds for that distractor ($x$), by
first checking whether the property exists in the STM
Buffer maintained by Consultant $c_x$ for $x$, or, if and only if this is
not the case, by checking whether the property is known to hold by
Consultant $c_x$ using its' $apply$ method, as usual.


\begin{figure}[ht]
\fbox{\parbox{\textwidth}{
\center{\textbf{Notation}}
\begin{scriptsize}
\begin{description}
\item [$C$] A set of \emph{Consultants} $\{c_0,\dots,c_n\}$
\item [$c_m^{\Lambda}$] The set of formulae
  $\{\lambda_0,\dots,\lambda_n\}$ advertised by Consultant $c$ responsible for $m$.
\item [$c_m^{\Lambda_{STM}}$] The STM buffer of formulae maintained by Consultant
  $c$ responsible for $m$.
\item [$M$] A robot's \emph{world model} of entities 
  $\{m_0\ldots m_n\}$ found in the domains provided by $C$. 
\item [$D$] The incrementally built up
  description, comprised of mappings from entities $M$ to sets of pairs
  $(\lambda,\Gamma)$ of formulae and bindings for those formulae.
\item [$D^M$] The set of entities $m\in M$ for which sub-descriptions
  have been created.
\item [$d^M$] The set of entities $m\in M$ involved in
  sub-description $d$. 
\item [$P$]  The set of candidate $(\lambda,\Gamma)$ pairs under
  consideration for inclusion.
\item [$Q$] The queue of referents which must be described.
\item [$X$] The incrementally pruned set of distractors
\end{description}
\end{scriptsize}
}}
\end{figure}

\begin{algorithm}[!ht]\centering\scriptsize
\caption{$\textit{SD-PIA}(m,C)$}
\begin{algorithmic}[1]
\STATE{$D =$ new Map() // \textit{The Description}} \label{alg:init1}
\STATE{$Q =$ new Queue($m$) // \textit{The Referent Queue}} \label{alg:init2}
\WHILE{$Q\neq\emptyset$} 
\STATE // \textit{Consider the next referent}
\STATE{$m\prime$ = pop(Q)}\label{alg:piapop}
\STATE// \textit{Craft a description $d$ for it}
\STATE{$(d,X)=\textit{SD-PIA-STM-H}(m\prime,C)$}\label{alg:pialoop1}
\STATE{$d=\textit{SD-PIA-H}(m\prime,C,X,d)$}\label{alg:pialoop1a}
\STATE{$D= D\cup \{m\rightarrow d\}$} 
\STATE// \textit{Find all entities used in $d$}
\FORALL{$m\prime\prime\in {d}^M\setminus keys(D)$}\label{alg:enqueue1}
\STATE // \textit{And add undescribed entities to the queue}
\STATE{$push(Q, m\prime\prime)$}\label{alg:enqueue2}
\ENDFOR \label{alg:pialoop2}
\ENDWHILE
\RETURN{$D$}\label{piaret}
\end{algorithmic}
\label{alg:distpia}
\end{algorithm}
\begin{algorithm}[!ht]\centering\scriptsize
\caption{$\textit{SD-PIA-STM-H}(m,C)$}
\begin{algorithmic}[1]
\STATE{$d = \emptyset$ // \textit{The Sub-Description}}
\STATE{$X = M \setminus m$ // \textit{The Distractors}}\label{alg:stmh1-1}
\STATE{$P = order([\forall \lambda \in c_m^{\Lambda_{STM}}:
    (\lambda,\emptyset)], c_m^{\Lambda})$}\label{alg:stmh1-2}
\WHILE{$X\neq\emptyset$  and  $P\neq\emptyset$} 
\STATE{$(\lambda,\Gamma)= pop(P)$}
\STATE{$\bar{X} = [x\in X \mid
    stm\_apply(c_x,\lambda,rebind(\Gamma,m\rightarrow x)) < \tau_{dph}]$}
\IF{$\bar{X}\neq\emptyset$}
\STATE{$d= d\cup (\lambda,\Gamma)$}\label{alg:stmh2}
\STATE{$X= X\setminus \bar{X}$}\label{alg:stmh3}
\ENDIF
\ENDWHILE
\RETURN{$(d,X)$}\label{alg:piah-ret}
\end{algorithmic}
\label{alg:stmh}
\end{algorithm}
\begin{algorithm}[!ht]\centering\scriptsize
\caption{$\textit{SD-PIA-H}(m,C,X,d)$}
\begin{algorithmic}[1]
\STATE{// \emph{Initialize a set of properties to consider: those
    advertised by the Consultant $c$ responsible for
    $m$ and not already part of the sub-description}}
\STATE{$P = [\forall \lambda \in c_m^{\Lambda}:
    (\lambda,\emptyset)] \setminus d$ }\label{alg:hinit2}
\STATE{// \emph{While there are distractors to eliminate or properties to consider}}
\WHILE{$X\neq\emptyset$  and  $P\neq\emptyset$} \label{alg:piah-startloop}
\STATE{$(\lambda,\Gamma)= pop(P)$}\label{alg:piahpop}
\STATE{// \emph{Find all unbound variables in the next property}}
\STATE{$V=find\_unbound(\lambda,\Gamma)$}\label{alg:piah-findunbound}
\IF{$\mid\!V\!\mid > 1$}\label{alg:piah-if1}
    \STATE{// \emph{If there's more than one, create copies under all possible variable bindings that leave one variable of the same type as the target unbound}}
\FORALL{$\Gamma\prime \in
  cross\_bindings(\lambda,\Gamma,C)$}\label{alg:piah-crossbindings}
\STATE{// \emph{And push them onto the property list}}
\STATE{$push(P,(\lambda,\Gamma\prime))$}\label{alg:piah-pushP}
\ENDFOR\label{alg:piah-if2}
    \STATE{// \emph{Otherwise, if sufficiently probable that the
    property applies to the target...}}
\ELSIF{$apply(c_m,\lambda,\Gamma\cup (v_0\rightarrow m)) >
  \tau_{dph}$}\label{alg:piah-else1}
    \STATE{// \emph{And sufficiently probable that it does
        \textbf{not} apply to at least one distractor...}}
\STATE{$\bar{X} = [x\in X \mid apply(c_x,\lambda,\Gamma\cup
    (v_0\rightarrow x)) < \tau_{dph}]$}
    \STATE{// \emph{Then bind its free variable to the target, and add it to the sub-description...}}
\IF{$\bar{X}\neq\emptyset$}
    \STATE{// \emph{And remove any eliminated distractors}}
\STATE{$d= d\cup (\lambda,\Gamma\cup (v_0\rightarrow m))$}
\STATE{$X= X\setminus \bar{X}$}\label{alg:assess2}
\ENDIF
\ENDIF\label{alg:piah-else2}
\ENDWHILE
\RETURN{$d$}\label{alg:piah-ret}
\end{algorithmic}
\label{alg:distpiahelper}
\end{algorithm}

\section{Demonstration}

We will now present a proof-of-concept demonstration of our
proposed algorithm, implemented in the ADE~\cite{scheutz2006ade} implementation of the DIARC
architecture~\cite{scheutz2018cognitive,schermerhorn2006diarc}.
The
ADE (Agent Development Environment) middleware provides a well-validated infrastructure for enabling
agent architectures through parallel distributed processing.
The
Distributed Integrated Affect 
Reflection Cognition (DIARC) Architecture is a component-based
architecture that has been under development for over 15 years, 
with a focus on robust spoken language
understanding and generation. For our demonstration scenario, the
following architectural components were used:
Speech Recognition (using the Sphinx4 Speech
Recognizer~\cite{walker2004sphinx}), Parsing (which uses the most recent
iteration~\cite{scheutz2017spoken} of the TLDL Parser~\cite{dzifcaketal09icra}), the
Dialogue and Pragmatics
Components~\cite{briggs2013hybrid},
the NLG
Component (in which the \sdp algorithm was implemented), the Goal Manager~\cite{schermerhorn2009utility}, the
Belief Component (which provides a Prolog Knowledge Base~\cite{clocksin2003programming}), the
Resolver Component~\cite{williams2016aaai},
the GROWLER HyperResolver Component~\cite{williams2018mrhrc:growler},
and a simulated Vision
Component~\cite{krauseetal14aaai} 
(which
serves as a Consultant).

For this demonstration walkthrough, we use a simple scenario involving
a single ``objects Consultant'' (the aforementioned simulated Vision
Component), which advertises a variety of constraints, e.g.,
related to object type and object color, with
type constraints having higher preference than color constraints and ordered according to specificity. The scene in
front of the robot contains a red teabox (known to the objects
Consultant as $object_1$) and a green teabox (known to the objects
Consultant as $object_2$) sitting on
a table (known to the objects
Consultant as $object_3$).


For this walkthrough, we begin by instructing the simulated robot
``Look at the box''.
The TLDL Parser~\cite{scheutz2017spoken,dzifcaketal09icra} parses this
into an utterance of type \emph{Instruction}, with primary semantics
$lookat(self,X)$ and supplemental semantics $\{box(X)\}$.
The GROWLER reference resolution algorithm
described in our recent work~\cite{williams2018mrhrc:growler} (see
also~\cite{williams2016hri,williams2018oxford}), then identifies
the two teaboxes $(object_1, object_2)$, as candidate referents
satisfying the given description. During the reference resolution process, when the
property $box(X)$ is determined to hold for each entity, it is placed
into that entity's STM Buffer within the simulated
Vision Component. Because the expression is ambiguous, a clarification
request is automatically generated~\cite{williams2017rss} to determine
whether $object_1$ or $object_2$ is the target referent. For each of
these entities, \sdp is recruited to generate referring
expressions. In this section, we will describe the process followed in
the selection of properties for $object_1$ alone (hereafter $o_1$), as
the process for $object_2$ is identical in structure.

\sdp begins by creating empty description $D = \emptyset$ and
referent queue 
$Q =\{o_1\}$ (Alg.~\ref{alg:distpia}, Lines~\ref{alg:init1}-~\ref{alg:init2}). 
Because there are still referents to describe (Line~\ref{alg:piapop}), \sdp calls helper
function \emph{SD-PIA-STM-H} (\emph{STM-H} hereafter) to craft a sub-description for $o_1$, which is popped
off of $Q$ (Line~\ref{alg:pialoop1}).

\emph{STM-H} begins by asking the Consultant responsible for $o_1$ for a set of 
distractors $X$ (e.g., $\{o_2,o_3\}$) 
and the set of properties $P$ stored in its STM Buffer
for $o_1$, sorted by that Consultant's preference ordering
$c^{\Lambda}_{m}$. 
(Alg.~\ref{alg:stmh} Lines~\ref{alg:stmh1-1}-\ref{alg:stmh1-2}), in this
case $box(X)$.
%
Next, \emph{STM-H} constructs the reduced set of distractors
$\bar{X}$ for which the property does \emph{not} appear in STM (i.e.,
$o_3$). Because this is nonempty, $o_3$ is removed from the set of
distractors, 
and $box(o_1)$ is added
to sub-description $d$ 
(Alg.~\ref{alg:stmh} Lines~\ref{alg:stmh2}-~\ref{alg:stmh3}).
%
Because there are no more properties to examine in $P$,
sub-description $d$ and the set of remaining distractors $\{o_2\}$ is
returned.

The second helper function, \emph{SD-PIA-H} (Alg.~\ref{alg:distpiahelper},
hereafter simply \emph{HELPER}), is
then used to complete the referring expression. \emph{HELPER}
begins by asking the Consultant responsible for $o_1$ for the set of
properties $P$ it can use in descriptions, sorted according to its predetermined
preference ordering, and ignoring properties already contained in
sub-description $d$, e.g.,
{teabox(X), table(X), red(X), green(X), on(X,Y)} 
(Alg.~\ref{alg:distpiahelper} Line~\ref{alg:hinit2}).

From this list, \emph{HELPER} pops the first unconsidered property
(i.e., $teabox(X)$) and its (empty) set of bindings. 
$teabox(X)$ has
exactly one unbound variable, so \emph{HELPER} will use 
Consultant $o\text{'s}$ $apply$ method (as per Capability 3) 
to ask how probable it is that $teabox(X)$ applies to $o_1$
(Alg.~\ref{alg:distpiahelper} Line~\ref{alg:piah-else1}).
Because it is sufficiently probable that this property applies to this
entity, 
\emph{HELPER} uses the same method to determine whether it also
applies to the single remaining distractor ($o_2$). Because it does,
the property will be ignored. 

\emph{HELPER} will then repeat this process with other properties.
Suppose it is insufficiently probable that $table(X)$ holds: 
it will be ignored.
Suppose it \emph{is} sufficiently probable that $red(X)$ holds
but not for the lone
remaining distractor ($o_2$), allowing $o_2$ to be ruled out.
Thus, $\{o_2\}$ will be
removed from $X$,
and
$red(o_1)$ will
be added to  $d$.
Since $X$ is now empty, 
the sub-description $\{(box(o_1), red(o_1)\}$ will be
returned to \sdp (Line~\ref{alg:piah-ret}) and added to
full description D. 
Since this
sub-description does not refer to any entities that have yet to be
described and $Q$ is empty, \sdp will
return description $D$
(Algorithm~\ref{alg:distpia}, Line~\ref{piaret}).
This process is then repeated for object $o_2$.
It will be the responsibility of the
next component of the natural language pipeline to translate this into
an RE, e.g. ``Do you mean the red box or the green box?''~\cite{williams2017rss}.

\section{Discussion}

\textbf{Potential Benefits:}
The primary motivation behind
our approach is performance: the number of queries needed
when determining what properties to use may be much lower
when those properties are sufficiently
discriminating. Similarly, determining whether properties
rule out distractors may be possible using set-membership checks
rather than costly long term memory queries.
Moreover, we believe these buffers may
facilitate \emph{lexical
  entrainment}~\cite{brennan1996lexical,brennan1996conceptual}, where
conversational partners converge on common choices of labels and
properties over the course of a conversation, resulting in more 
comprehensible referring expressions~\cite{tolins2017overhearing}.
If a robot's STM Buffers are populated with properties used by itself and
its interlocutors, and if the properties contained in those buffers
are considered before others, this may directly lead to such entrainment.

\noindent\textbf{Potential Limitations:}
On the other hand, because the robot arbitrarily restricts itself to a
subset of the properties it \emph{could} otherwise choose to use, it
may force the robot into local maxima in the landscape of 
referring expressions. Moreover, the robot runs the risk of using a
property that does not actually hold if it does not appropriately
handle contextual dynamics. For example, an object previously
described as ``on the left'' may no longer be ``on the left'' if the
object, robot, or interlocutor has moved since the object was last
discussed.

\noindent\textbf{Design Decisions: Buffer Size Limitations:}
Many insights from psychology could be leveraged to prevent such mistakes. A
context-sensitive decay-based model of working memory might prevent
this by having different properties ``decay'' out of cache, with time or
probability of decay proportional to how
likely it is to change over
time~\cite{baddeley1975word,schweickert1986short}. A resource-based
model might prevent this by having a limited total buffer size, and
have property dynamics factor into the decision of what to bump from
memory when new things need to be inserted into an already-full
buffer~\cite{just1992capacity,ma2014changing}.
Finally,
an interference-based model might prevent this by having
properties added to a buffer ``overwrite'' the most similar property
currently in the
buffer~\cite{oberauer2006formal,saito2004nature}. 
These are loose characterizations of the respective theories
from cognitive psychology; a comprehensive discussion of these
theories and the relative evidence for them from a psychological
perspective can be found in~\cite{oberauer2016limits}.
Of course, the approach taken need not be cognitively plausible. The
robot could, for example, statistically model the dynamics of
different properties, and use them to periodically re-sample the
properties held in its buffers.

The question of cognitive plausibility also raises a different
question: how extensive should the robot's memory caches be? Should
the robot keep property caches for all entities, for only those that
are relevant in the current context, or for an even smaller set? And
for each entity, should the robot track all relevant knowledge for so
long as that entity is tracked, or should it track only a fixed, small
number of properties? And should such limits be local, or global
limits shared between tracked entities? These are once again questions
for which candidate answers can be gleaned from the psychological
literature~\cite{oberauer2016limits}. Here, interesting tradeoffs can
be made. On the one hand, robots can be made to remember much more
than humans can. On the other hand, expanded memory may come at a
computational cost; and moreover, choosing to remember more means
increased risk of incorrect behavior due to mishandling of property
dynamics. Further evidence from neuroimaging studies suggests that the
contents of working memory in humans is biased by humans' current
goals, with only task-relevant features being maintained in visual working
memory~\cite{yu2017occipital}. For robots, task-relevance may be
similarly useful for optimally selecting what to maintain in
Consultants' STM Buffers.

\noindent\textbf{Design Decisions: Buffer Population:}
Similar tradeoffs arise when deciding when to add properties
used in natural language to a robot's STM Buffers. In this work,
properties are added to buffers as soon as they are determined to hold
for a referent, 
rather than
waiting until the end of the reference resolution process.
This 
means that lexical entrainment effects may be seen for
entities other than intended referents. For example, consider 
``the tall red box''. Assuming properties are
processed in the order $tall(X)$, $red(x)$, $box(X)$, all tall
objects in the scene will have $tall(X)$ added to their STM Buffers,
all tall red objects will also have $red(X)$ added to their STM
Buffers, and tall red boxes will have all three properties added to
their STM Buffers. Accordingly, the robot will prefer to use height
and potentially color and shape to refer to other objects
even if they have not been referred to. This is not dissimilar from
psycholinguistic observations
of syntactic and semantic carry-over
from previous object references to 
references to previously
unmentioned
entities~\cite{goudbeek2012alignment,carbary2011conceptual}.
This could be particularly effective in the 
case of the Vision Consultant, as additional caching reduces the risk
of needing to conduct future (potentially expensive) visual
searches. If this particular memory-performance tradeoff is not
optimal for Consultants associated with non-visual modalities, the
decision to add properties to STM Buffers may need to be made on a
per-Consultant basis.

\section{Conclusions}

We have presented a caching strategy augmenting a
robot's set of distributed knowledge sources with cognitively
inspired STM Buffers so as to increase computational
efficiency. We have explained how these buffers may facilitate linguistic
phenomena such as lexical entrainment, and identified insights from
cognitive science that may inform future
refinements of these Buffers. Two tasks will be
crucial for future research building off this work. First, the
presented approach must be evaluated both in terms of computational
performance and with respect to objective and subjective
measures such as those used in our previously presented evaluative
approach~\cite{williams2017inlg}. Second, insights from cognitive
science must be leveraged to enable alternate design decisions that may be explored
both from the perspectives of artificial intelligence for human-robot
interaction, cognitive modeling,
and cognitive architecture.
\vspace*{-3mm}
\subsection*{Acknowledgments}
\vspace*{-2mm}
\begin{small}
This work was in part supported by ONR grant N00014-16-1-0278.
\end{small}
\vspace*{-3mm}
\bibliographystyle{splncs04}
\def\bibfont{\footnotesize}
\bibliography{references}
\end{document}